\documentclass[graybox]{svmult}

\usepackage{mathptmx}       
\usepackage{helvet}         
\usepackage{courier}        
\usepackage{type1cm}        
\usepackage{makeidx}         
\usepackage{graphicx}        
\usepackage{multicol}        
\usepackage[bottom]{footmisc}
\usepackage{float}

\makeindex             

\begin{document}

\title*{Wayfinding and cognitive maps for pedestrian models}
\author{Erik Andresen, David Haensel, Mohcine Chraibi, and Armin Seyfried}
\authorrunning{E. Andresen et al.}
\institute{Erik Andresen \at Faculty of Architecture and Civil Engineering, Bergische Universit\"at Wuppertal, Pauluskirchstr. 7, 42285 Wuppertal, Germany \email{andresen@uni-wuppertal.de}
\and David Haensel \at J\"ulich Supercomputing Center, Forschungszentrum J\"ulich GmbH, 52425 J\"ulich, Germany \email{d.haensel@fz-juelich.de}
\and Mohcine Chraibi \at J\"ulich Supercomputing Center, Forschungszentrum J\"ulich GmbH, 52425 J\"ulich, Germany \email{m.chraibi@fz-juelich.de}
\and Armin Seyfried \at J\"ulich Supercomputing Center, Forschungszentrum J\"ulich GmbH, 52425 J\"ulich, Germany\\
Faculty of Architecture and Civil Engineering, Bergische Universit\"at Wuppertal, Pauluskirchstr. 7, 42285 Wuppertal, Germany \email{a.seyfried@fz-juelich.de}}
%
%
\maketitle

\abstract{Usually, routing models in pedestrian dynamics assume that agents have fulfilled and global knowledge about the building's structure. However, they neglect the fact that pedestrians possess no or only parts of information about their position relative to final exits and possible routes leading to them. To get a more realistic description we introduce the systematics of gathering and using spatial knowledge.  
A new wayfinding model for pedestrian dynamics is proposed. 
The model defines for every pedestrian an individual knowledge representation implying inaccuracies and uncertainties.
In addition, knowledge-driven search strategies are introduced.
The presented concept is tested on a fictive example scenario.}

\section{Introduction}
\label{introduction}

Microscopic simulations of pedestrian traffic flow are a suitable tool for designing both escape routes in buildings and pedestrian areas, e.g. malls, train and bus stations, etc.. Besides, simulations are used to investigate and analyze security risks in advance.

In the literature many elaborated microscopic pedestrian traffic flow models can be found. For a first overview see \cite{Schadschneider.2009}. These models try to describe the locomotive actions of pedestrians, e.g. basic movement towards a certain location in space or steering (around obstacles to a certain destination).

However, they neither include the choice between currently accessible targets nor the planning of proceeding destinations (wayfinding tasks). These tasks are covered by the tactical level of pedestrian traffic flow modeling (see for example \cite{KemlohWagoumArmelUlrich.2013, Crociani.2014}). 

A majority of models concerning the tactical level assume the pedestrians to have a comprehensive knowledge about the spatial structure of their environment. Thus, the agents possess the ability to localize desired destinations in advance. They are, further more, able to evaluate or rather compare the quality of the routes which lead to the destinations. In many cases their evaluations are based on shortest path calculations or travel time optimization.

The assumption that all pedestrians are provided with comprehensive global knowledge about a building's structure is a rough approximation, for example when pedestrians are not familiar with the facility. Even less, they are able to evaluate metric information about multiple routes so that an exact comparison is possible. In fact, the knowledge status of a group of pedestrians vary according to the number of visits and the capability to learn the spatial structure of new environments. 

Human wayfinding is a complex process which includes the use of (in some cases inaccurate and incomplete) spatial memories \cite{Wiener.2009} , the use of signs and maps \cite{Wiener.2009}, search strategies and herding phenomena.  

Although there are already approaches to represent wayfinding aspects including directional knowledge and uncertainties \cite{Kneidl.2013} there is room for improvements and continuations. 

In this work we introduce a modeling approach enabling agents to make exit choice decisions based on inaccurate and incomplete knowledge about their environment and destinations.

\subsection{The cognitive map}
\label{subsec:2}
Although many mechanisms of perception and cognition enabling successful wayfinding are still unacquainted it is known that the hippocampal formation (part of the limbic system of the human brain) is mainly responsible to store and retrieve spatial memories which are essential to solve wayfinding issues \cite{OKeefe.1978}. John O'Keefe \cite{OKeefe.1978} and Maybritt and Edvard Moser \cite{Moser.2008} 
discovered place cells and grid cells in rats' brains that are involved in the formation of the so-called cognitive map. Similar systems of place-like and grid-like cells were discovered in many mammals' brains including the human brain \cite{Ekstrom.2003}.
 
The term cognitive map has been introduced by Tolman \cite{Tolman.1948}. It depicts the mental representation of the spatial relationships between essential points, places, objects, etc. of our environment and possible connections between them \cite{Golledge.2000}. Despite the prevailing opinion rats can only respond to stimuli Tolman \cite{Tolman.1948} conducted some experiments which gave evidence about the fact that rats possess clues about specific objects' positions relative to each other gathered from previous visits of the environment. 

In the best case the cognitive map provides the possibility to locate the relative position to a specific destination and enables us to find or to plan a route leading to this destination \cite{Golledge.2000, Ellard.2009}.

However, there is evidence that people get lost in several situations due to the fact that their cognitive maps are inaccurate, incomplete, distorted, or even wrong \cite{Golledge.2000, Ellard.2009}.

Nevertheless, the cognitive map, although it does not provide detailed and much less accurate metric information, successfully helps us to find our way in most situations, especially in environments visited multiple times before. This results from the fact that humans possess the ability to store topological relations in a more accurate way \cite{Golledge.2000, Ellard.2009}. 

\subsection{Generalized knowledge}
\label{generalized}
In many cases a wayfinding problem is not merely solved by information about the relations of explicit points or objects (the cognitive map). Additional knowledge called generalized knowledge is used as well. 
Human beings classify their environments and retrieve information, implications and expectations about the according classes, for example train stations, libraries, office buildings, etc. \cite{Anderson.2010, St.Pierre.2014}. Generalized knowledge does not concern the explicit set-up of the specific environment itself but information about the environment's type or rather classification. 

Within buildings we differentiate between two types of rooms (enclosed areas). On the one hand there are rooms serving the building's circulation or rather enabling people to reach efficiently their destination areas. Corridors, entrances, lobbies, stairs, ramps, etc. belong to this group of rooms. On the other hand there are rooms allotted to an explicit usage excluding the circulation. Concerning the second type of rooms we mention functional rooms, common rooms (offices, living rooms, cafeterias, etc.), store rooms, etc. as examples. We assume the majority of people to be capable to distinguish between both mentioned types due to their generalized knowledge about spatial structures.

Generalized knowledge provides the basis for various search strategies. 
To mention an example we consider a person to be located somewhere in a completely unfamiliar office building. The person is going to leave the building and is therefore looking for an exit. Due to knowledge about the purpose of circulation rooms he/she prefers to use them to reach the exit. 
Preferring circulation rooms instead of others is a simple but expedient and efficient strategy compared to a simple room exploration and thus facilitates the search for the exit severely.

\section{Modeling cognitive map knowledge}
\label{explicitk}

\begin{figure}[!hbtp]
 \includegraphics[trim = 0 0 0 50, width=\textwidth]{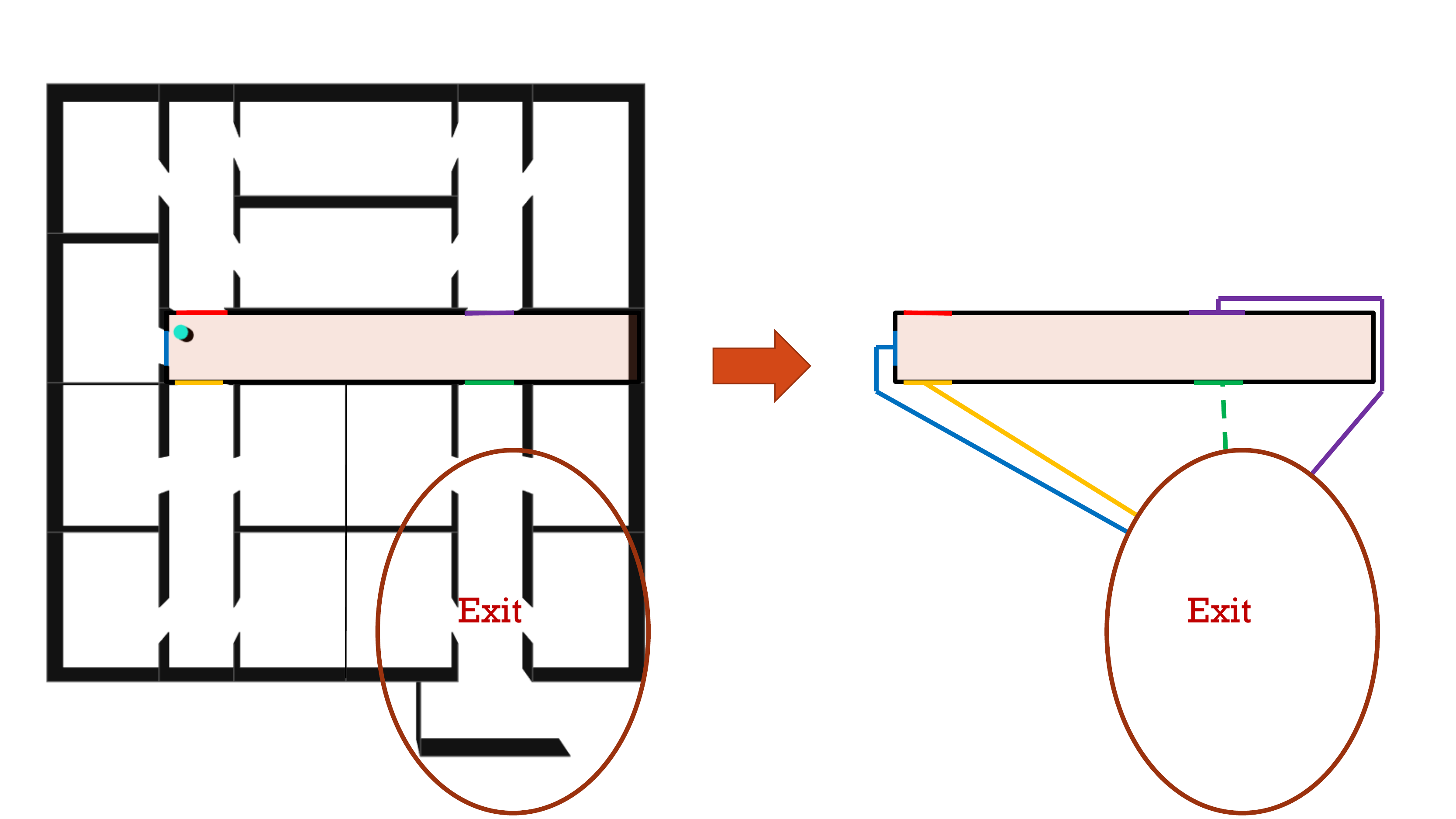}
 \caption{On the left: The figure shows the layout of a fictional building's floor. The agent's actual room is highlighted. The ellipse represents the agent's inaccurate idea of the exit's position.
On the right: The figure shows the agent's actual room and the ellipse without exception. The lines indicate possible routes to the exit region ignoring the spatial structure beyond the actual room.  
The shortest path from a doorway of the room to the exit area is shown dashed.}
 \label{tgfscenario2}
 \end{figure}

Following the findings mentioned in Sec. \ref{subsec:2} we assume a simulated person to possess a cognitive map consisting of uncertain, inaccurate information. Thus, the agents possess only a vage idea of the exact (sub-)goals' position. For this purpose the inaccurate memories of the goals are not restricted to a point location but are represented by ellipses (see Fig. \ref{tgfscenario2}).      

We assume that the agent searches a route leading him, preferably following the beeline, to the exit area. Therefore he chooses a doorway leading him as closely as possible to the destination area. Due to the fact that the agent has no knowledge about the remaining structure of the building his decision is only made by considering position and shape of the actual room (and its doors) and the ellipse representing the approximate position of the exit. Further rooms or rather their walls or obstacles beyond the actual room are not familiar to the agent. Hence, they are not taken into account within the decision making.  
 
To determine the doorway which takes the agent as closely as possible to the ellipse the (lengths of the) shortest paths between every accessible doorway and the ellipse are calculated (see Fig. \ref{tgfscenario2}). The shortest path calculation is only performed under the consideration of the current room's walls (as obstacles). Even if the made assumption of an empty area beyond the current room may be inaccurate in most cases, this procedure will find the most appropriate doorway to come closer to the exit area if no spatial information of proceeding areas is available.      

Depending on the length of their paths compared to paths from other possibilities the doorways are weighted differently. The doorway related to the shortest one of all shortest paths will be preferred by the pedestrian (see Fig. \ref{tgfscenario2}, right, dashed line).

If the agent has arrived at the target area (is located inside the ellipse) and there is still no exit in sight he has to rely on other information or strategies to look for the continuative way to the exit.

\section{Examples}

In the next section we demonstrate how the model presented performs in simple scenarios with respect to different degrees of spatial knowledge. 

\begin{figure}[!hbtp]
 \includegraphics[trim= 100 20 100 20, width=\textwidth]{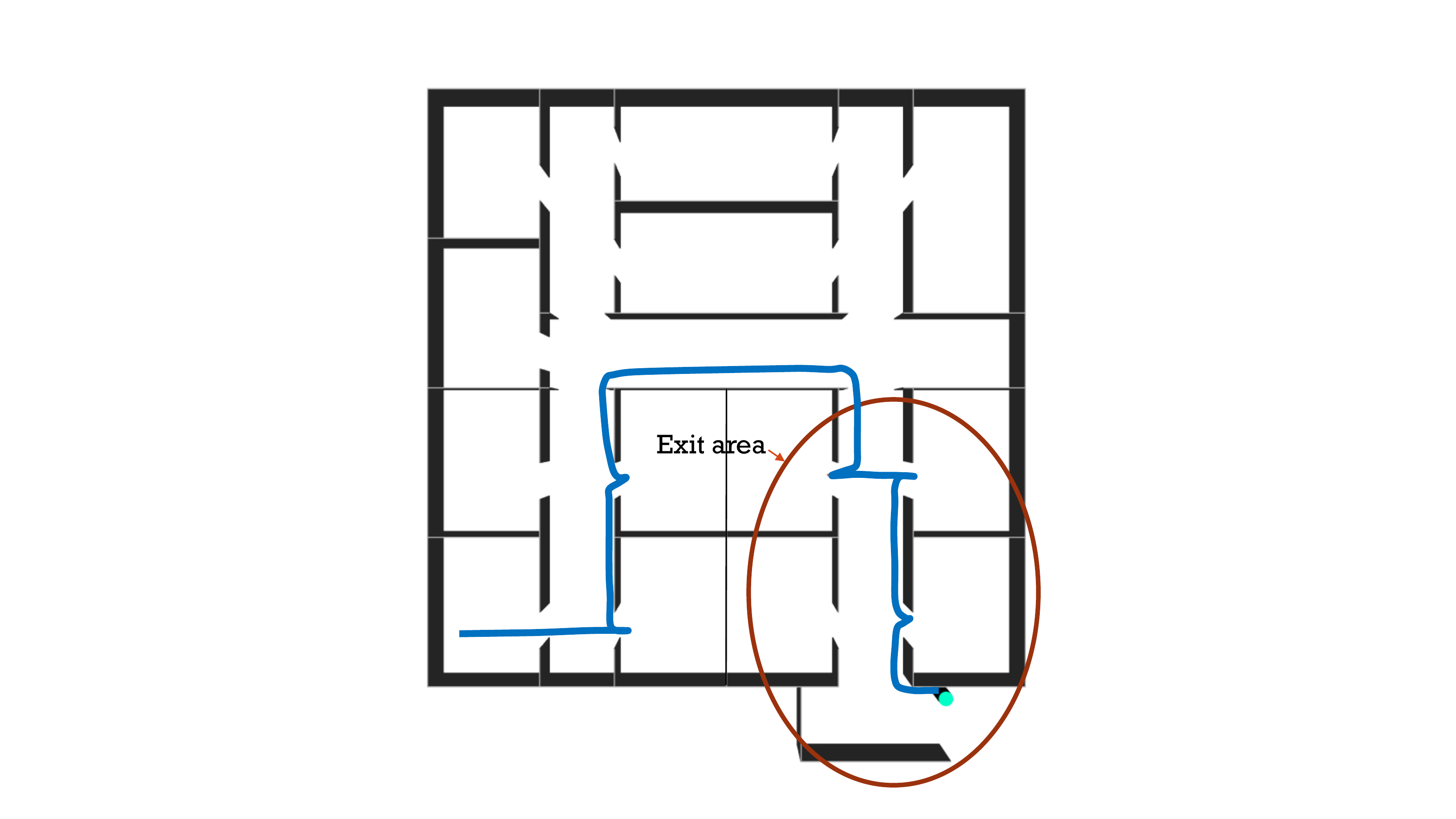}
 \centering
 \caption{Trajectories of an agent situated in a fictional office building. The agent has access to cognitive map knowledge depicted with the help of the ellipse.}
 \label{cognitivemaplike}
 \end{figure}

In every scenario an agent is situated in a room at the left lower corner of a fictional building (see Fig. \ref{cognitivemaplike}) and searches a way to the outside. The entrance / exit can be found at the right lower corner.
 
\subsection{Scenario 1: Cognitive map knowledge}

With the help of this scenario we investigate the effects of using the modeling approaches concerning explicit cognitive map knowledge (see. Sec. \ref{explicitk}). The agent supposes the exit to be somewhere in the area depicted by the ellipse in the right lower region. At every choice point he decides to move to the direction taken him closer to the assumed area location of the exit.

Having started his journey the agent crosses the first corridor (lower left corner) heading to the opposing doorway as it is obviously the best choice to come closer to the exit area, assuming the lack of knowledge about the structure beyond the doors.  
However, as the agent recognizes that he is located in a dead end he turns around trying to reach the exit area by moving through the crossing to the adjacent room.  
Eventually, he arrives at the corridor located in the middle of the building which enables him to travel to the right lower region of the building.
Inside the ellipse depicting the exit area the agent proceeds to find the exit by exploring the rooms in the surrounding. He starts by heading to the nearest doorway.
After having explored three further rooms within the exit area the agent finally reaches the exit.

\subsection{Scenario 2: Combination of generalized and cognitive map knowledge}
\label{combinationscenario}

Scenario 2 comprises the combination of generalized and cognitive map knowledge. To highlight the effects of this combination the agent is simultaneously provided with the ability to distinguish between common rooms and circulation rooms and with a directional sense of the exit's location.
For this purpose the color coded rooms are indicated as circulation rooms. Doorways leading to these rooms will be preferred by the agent. Further more, the agent is following the procedure explained in Sec. \ref{explicitk}. 

\begin{figure}[!hbtp]
 \includegraphics[trim= 100 10 100 10, width=\textwidth]{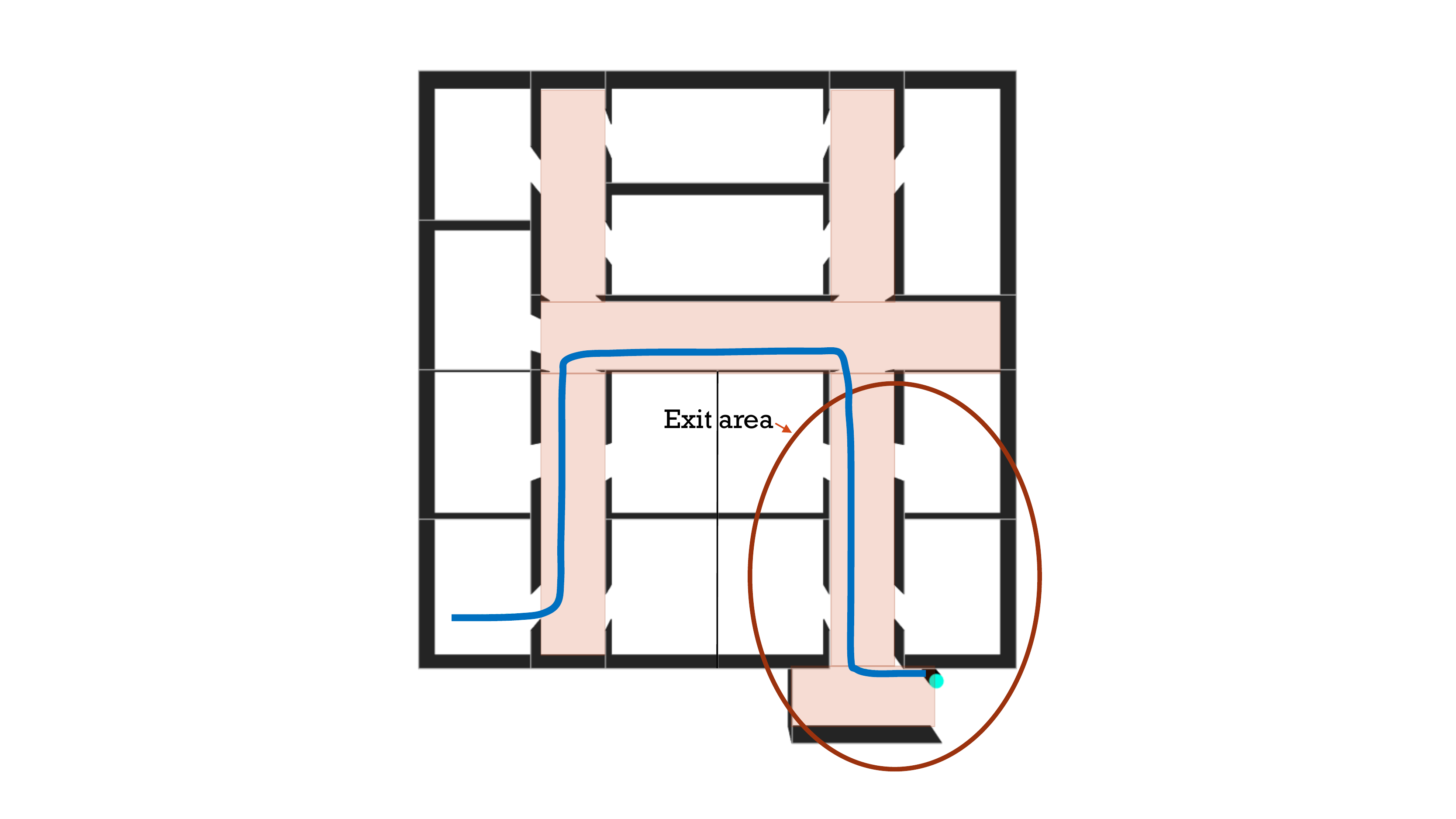}
 \centering
 \caption{Trajectories of an agent situated in a fictional building. The agent has access to both cognitive map knowledge and generalized knowledge.}
 \label{combination}
 \end{figure}

We assume the strategy to move to or to stay on a circulation room to be more expedient as to keep the direction to the destination. Following this assumption, the pedestrian will use circulation rooms even though he has to depart seriously from the beeline to the exit area.

Leaving the starting room the agent proceeds to the corridor in the middle of the building as it is the only adjacent corridor. 
Being located in the middle corridor the agent has to choose between three corridors. 
Obviously, the corridor in the right lower corner is the best possibility to come closer to the exit area. 
Within the exit area the agent again prefers the only proceeding corridor taking him eventually to the outside.  

In this example scenario the agent is moving to the destination without making any detours. Hence, the search strategy (go to and stay on circulation rooms) and a vage idea about the location of the destination are sufficient in this example case.

\section{Summary and Outlook}

The modeling approaches introduced in this paper provide simulated agents with restricted information about their environment instead of granting them access to global comprehensive knowledge about every part of the environment's structure. Additionally, the restricted information consists of uncertainties and inaccuracies. The information status of agents can be manipulated by modifying position and size or shape of ellipses modeling its actual cognitive map. In addition, it is conceivable to vary the knowledge degree of an agent compared to other agents by differing modifications.
 
Based on two examples we demonstrated the effects of different knowledge degrees. The first example showed that the agent does not instantly find an appropriate route to the outside by simply heading to the exit area.  
In the second example (Sec. \ref{combinationscenario}) it has been shown that a vage, inaccurate idea of the destination's location in combination with the use of a search strategy is sufficient to find a route leading directly (without detours)  
to the desired destination.

Proceeding work implies the creation of a continuative framework modeling the human wayfinding process. On the one hand the framework is supposed to contain further mechanisms of the cognitive map, for example the involvement of landmarks and self localization procedures. On the other hand it is supposed to include search strategies, recognition of signs and herding effects. 

The affiliation of further models representing factors which contribute to exit choice decisions beside the wayfinding process is possible. Concerning further factors we mention sensory input models according to the evaluation of congestions \cite{KemlohWagoumArmelUlrich.2013} and smoke propagation \cite{Schroeder.2015}. 

\begin{acknowledgement}
This research is founded by the Deutsche Forschungsgemeinschaft (DFG) contract No. GZ: SE 17894-1.
\end{acknowledgement}

\bibliographystyle{spmpsci}
\bibliography{tgf2015}

\end{document}